\documentclass[sigconf]{acmart}

\usepackage{amsmath,amsfonts,bm}

\def\eqref#1{equation~\ref{#1}}

\def\1{\bm{1}}

\def\ve{{\bm{e}}}

\def\vp{{\bm{p}}}

\def\vx{{\bm{x}}}

\def\vz{{\bm{z}}}

\def\mE{{\bm{E}}}

\def\mP{{\bm{P}}}

\def\mZ{{\bm{Z}}}

\DeclareMathAlphabet{\mathsfit}{\encodingdefault}{\sfdefault}{m}{sl}
\SetMathAlphabet{\mathsfit}{bold}{\encodingdefault}{\sfdefault}{bx}{n}

\DeclareMathOperator*{\argmin}{arg\,min}

\usepackage{graphicx}
\usepackage{amsmath}
\usepackage{booktabs}
\usepackage{caption}
\usepackage{subfigure}
\usepackage{multirow}
\usepackage{wrapfig}
\usepackage{caption}
\usepackage{xspace}
\usepackage{colortbl}
\usepackage{makecell}
\usepackage{balance}
\usepackage{bbm}

\setcopyright{acmcopyright}
\copyrightyear{2018}
\acmYear{2018}
\acmDOI{XXXXXXX.XXXXXXX}
\begin{document}

\title{VPA: Fully Test-Time Visual Prompt Adaptation}

\author{Jiachen Sun}
\authornote{This project was mainly conducted during Jiachen Sun's internship at Meta AI.}
\affiliation{%
  \institution{University of Michigan}
  \city{Ann Arbor}
  \state{MI}
  \country{USA}
}
\email{jiachens@umich.edu}

\author{Mark Ibrahim}
\affiliation{%
  \institution{Meta AI}
  \city{New York}
  \state{NY}
  \country{USA}
}
\email{marksibrahim@meta.com} 

\author{Melissa Hall}
\affiliation{%
  \institution{Meta AI}
    \city{New York}
    \state{NY}
  \country{USA}
}
\email{melissahall@meta.com}

\author{Ivan Evtimov}
\affiliation{%
  \institution{Meta AI}
    \city{Seattle}
    \state{WA}
  \country{USA}
}
\email{ivanevtimov@meta.com}

\author{Z. Morley Mao}
\affiliation{%
  \institution{University of Michigan}
    \city{Ann Arbor}
  \state{MI}
  \country{USA}
}
\email{zmao@umich.edu}

\author{Cristian Canton Ferrer}
\affiliation{%
  \institution{Meta AI}
      \city{Seattle}
    \state{WA}
  \country{USA}
}
\email{ccanton@meta.com}

\author{Caner Hazirbas}
\affiliation{%
  \institution{Meta AI}
      \city{New York}
    \state{NY}
  \country{USA}
}
\email{hazirbas@meta.com}







\renewcommand{\shortauthors}{Jiachen Sun et al.}

\begin{abstract}
  Textual prompt tuning has demonstrated significant performance improvements in adapting natural language processing models to a variety of downstream tasks by treating hand-engineered prompts as trainable parameters. Inspired by the success of textual prompting, several studies have investigated the efficacy of visual prompt tuning. In this work, we present Visual Prompt Adaptation (VPA), the first framework that generalizes visual prompting with test-time adaptation. VPA introduces a small number of learnable tokens, enabling fully test-time and storage-efficient adaptation without necessitating source-domain information. We examine our VPA design under diverse adaptation settings, encompassing single-image, batched-image, and pseudo-label adaptation. We evaluate VPA on multiple tasks, including out-of-distribution (OOD) generalization, corruption robustness, and domain adaptation. Experimental results reveal that VPA effectively enhances OOD generalization by 3.3\% across various models, surpassing previous test-time approaches. Furthermore, we show that VPA improves corruption robustness by 6.5\% compared to strong baselines. Finally, we demonstrate that VPA also boosts domain adaptation performance by relatively 5.2\%. Our VPA also exhibits marked effectiveness in improving the robustness of zero-shot recognition for vision-language models.
\end{abstract}





\maketitle

\section{Introduction}
\label{sec:intro}

Visual recognition, a crucial component in multimedia systems, plays an essential role in various applications. As technology evolves and the demand for intelligent multimedia systems increases, the importance of effective and robust visual recognition techniques cannot be overstated.
Although various deep neural networks achieve state-of-the-art (SOTA) performance on test sets drawn from the same distribution as the training set~\cite{deng2009imagenet}, these expertly-trained models may struggle to generalize when faced with distribution shifts, leading to substantial performance drops~\cite{hendrycks2021natural,hendrycks2021many}. These shifts encompass common corruption~\cite{hendrycks2019benchmarking}, adversarial attacks~\cite{7958570,sun2023pointdp,sun2020adversarial,sun2021adversarially}, conceptual changes~\cite{hendrycks2021many,kim2021domain}, and even out-of-distribution (OOD)\footnote{In this study, we refer out-of-distribution (OOD) to \textit{covariate} shift but not \textit{concept} shift of the test/validation dataset.} variations~\cite{hendrycks2021natural}, and can emerge in numerous real-world applications such as autonomous driving~\cite{255240,zhang2022adversarial,sun2023calico} and facial recognition systems~\cite{vakhshiteh2021adversarial}, where accurate and robust performance is critical. Therefore, addressing the vulnerabilities to distribution shifts is essential for enhancing the robustness and generalization capabilities of deep learning models.

Numerous architectural improvements and training techniques have been proposed to address the challenges associated with achieving robustness against various domain variations~\cite{dosovitskiy2020image,radford2021learning,he2022masked}. For instance, recent advances in Vision Transformer (ViT) architectures~\cite{dosovitskiy2020image} have demonstrated significant improvements on many out-of-distribution (OOD) generalization and corruption robustness benchmarks~\cite{hendrycks2021natural,hendrycks2021many}. Pretraining and fine-tuning strategies, such as CLIP~\cite{radford2021learning} and WiSE~\cite{wortsman2022robust}, have further enhanced the generalization performance of ViT models~\cite{dosovitskiy2020image}. In addition to these general methods, a plethora of specialized training recipes have been presented to address specific objectives, including OOD generalization~\cite{wang2020tent,zhang2021memo}, corruption robustness~\cite{hendrycks2019augmix,cubuk2020randaugment}, domain adaptation~\cite{hendrycks2019augmix}. However, it remains challenging to address these generalization problems solely during the training phase, as a single training recipe cannot encompass all underlying distributions. Test-time updates serve as valuable complements, focusing on tailored adaptation for specific test data~\cite{wang2020tent}. Such a scheme is particularly important for multimedia systems, where content may come from unseen domains.

Humans typically begin with their existing knowledge and extrapolate from it when learning a new skill. Prompting is a similar paradigm that aids machine learning models in adapting to various contexts or even new tasks through specific textual input. This approach has gained popularity in the field of natural language processing (NLP)~\cite{liu2021pre}. Recent studies have shown that prompt tuning can enhance model generalization across different domains, where predefined prompts evolve as trainable parameters~\cite{zhou2022learning,zhou2022conditional}. Test-time Prompt Tuning (TPT) is a pioneering technique that leverages textual prompting during testing to improve the generalization capability of vision-language models~\cite{shu2022test,radford2021learning}. Prompt tuning is efficient in adapting a pretrained model, as it does not modify the original model parameters. In addition to prompting in NLP, recent studies have explored \textit{visual} prompting during the training phase~\cite{jia2022visual,bahng2022exploring}, yielding substantial improvements on numerous vision benchmarks. However, there is still a scarcity of research examining the application of visual prompting in online test-time adaptation, indicating an area ripe for further exploration.

\noindent\textbf{Our Contributions}. In this paper, we propose visual prompt adaptation (VPA) to bridge the gap between visual prompting and online test-time adaptation, drawing inspiration from the success of textual prompting in NLP. VPA is a simple yet effective framework that generalizes existing prompt designs and adaptation objectives. Given a pretrained model, we attach additive and prependitive prompts to the frozen model during the adaptation phase. VPA is highly storage-efficient. Rather than adapting all the parameters of the model, VPA requires only a small number of prompts to be stored. This efficiency greatly reduces the storage overhead while maintaining the ability to effectively adapt the model to new contexts and tasks, making VPA a practical and appealing solution for real-world applications.
In contrast to the pixel-space prompts in~\cite{bahng2022exploring} and the randomized initialization for embedding-space prompts in~\cite{jia2022visual}, we design a straightforward but intuitive paradigm using zero attention to initialize our prompts, ensuring that the original performance remains unaffected. We combine VPA with various adaptation settings, including batched-image adaptation (BIA), single-image adaptation (SIA), and pseudo-label adaptation (PLA). For BIA, we input a batch of images belonging to different classes into the model simultaneously and leverage self-entropy minimization as the adaptation objective. In SIA, we employ marginal entropy minimization as the objective by enriching a single image into a batch through various augmentations. Additionally, we employ confidence selection to identify images with high confidence, which allows for more effective adaptation. By focusing on these high-confidence images, the VPA framework can better leverage the information contained within them. Pseudo-labels have been shown to be effective in enhancing test-time adaptation performance. As such, we incorporate a memory queue that stores pseudo-labels for historical data to assist incoming images during adaptation. Our VPA combines the strengths of VPA, BIA, SIA, and PLA to achieve both \textit{fully} test-time and \textit{storage-efficient} adaptation.

We conduct extensive evaluations of VPA across three critical axes in real-world vision systems: OOD generalization (\S~\ref{sec:eval_ood}), corruption robustness (\S~\ref{sec:eval_corruption}), and domain adaptation (\S~\ref{sec:eval_da}). In our OOD generalization experiments, we are the first to evaluate visual prompting on a variety of large-scale datasets. We primarily use the ViT architecture fine-tuned on the ImageNet training set as our base model. Our results indicate that VPA enhances the average accuracy of ImageNet-scale OOD generalization benchmarks~\cite{hendrycks2021natural,hendrycks2021many} by 3.3\%. In contrast, existing state-of-the-art methods, such as TENT~\cite{wang2020tent} and DDA~\cite{gao2022back}, struggle to perform effectively under challenging OOD scenarios. Moreover, VPA achieves a similar 2.6\% improvement as MEMO~\cite{zhang2021memo} under SIA without updating the parameters of the frozen model. Notably, we also demonstrate that VPA enhances corruption robustness and domain adaptation performance by relative margins of 6.5\% and 5.2\%, respectively, compared to strong baselines~\cite{chen2022contrastive,kojima2022robustifying}. Moreover, we have shown that VPA could effectively improve the robustness of zero-shot recognition for vision-language models~\cite{radford2021learning}. It is important to note that the goal of our study is not solely to pursue state-of-the-art results but to highlight the potential applications of visual prompting in test-time adaptation. Our promising results will encourage future research to develop new adaptation schemes using visual prompting.

We summarize our main contributions as three-fold:

\noindent$\bullet$ We propose visual prompt adaptation (VPA), a \textit{fully} test-time and \textit{storage-efficient} adaptation framework that introduces both additive and prependitive adaptable tokens to improve the robustness of vision models.

\noindent$\bullet$ We conduct a rigorous taxonomy of VPA under different adaptation setups, including batched-image, single-image, and pseudo-label adaptation.

\noindent$\bullet$ We perform an extensive evaluation of VPA on various tasks, including out-of-distribution (OOD) generalization, corruption robustness, and domain adaptation. Our VPA consistently improves performance on these benchmarks by a significant margin.
\vspace{-0.4cm}
\section{Related Work}
\label{sec:related_work}
In this section, we review topics related to our study, including prompting in foundation models, test-time adaptation, and out-of-distribution (OOD) robustness and domain adaptation.  

\begin{figure*}[t]
  \centering 
  \includegraphics[width=\linewidth]{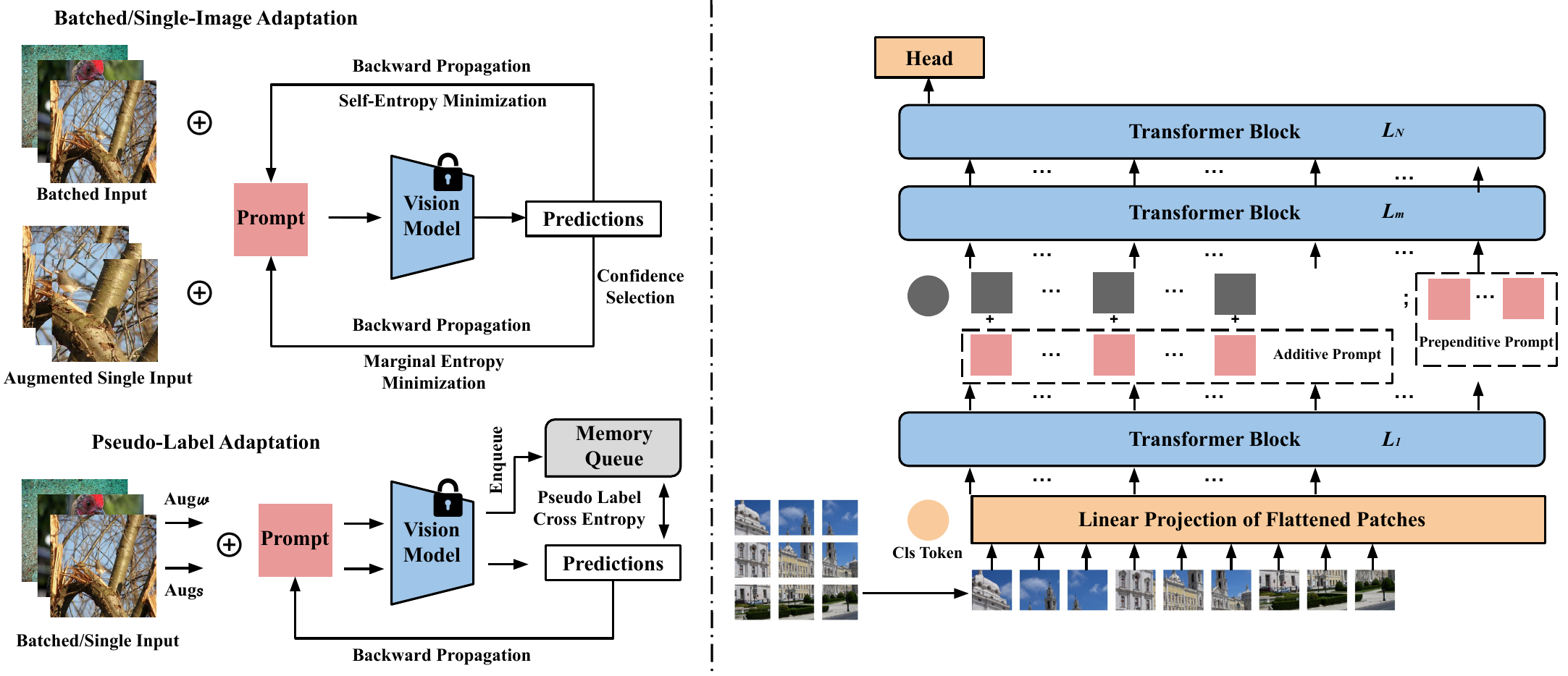}
  \vspace{-0.4cm}
  \caption{Overview of Visual Prompt Adaptation Pipeline. Our VPA supports both batched- and single-image and pseudo-label adaptation settings as shown in the left figure. The visual prompt designs are illustrated in the right figure based on the ViT architecture.}
  \label{fig:overview}
\end{figure*}

\noindent\textbf{Prompting in Foundation Models}. Deep learning has made significant strides in natural language processing and computer vision tasks~\cite{lecun1995convolutional,lecun2015deep}. In addition to architectural improvements~\cite{dosovitskiy2020image,he2016deep,vaswani2017attention}, recent efforts have focused on fine-tuning foundational models with large-scale data to enable transfer learning across multiple downstream tasks~\cite{devlin2018bert,radford2021learning}. Prompting, which originated in language models, involves using human-engineered texts to improve the context-specific learning of a given task~\cite{liu2021pre,kumar2016ask}. Prompt tuning has been advanced for different goals in language models~\cite{lester2021power,gu2021ppt,han2021ptr}, while visual prompting has been explored for zero-shot recognition tasks~\cite{zhou2022learning,zhou2022conditional}. For instance, CoOp~\cite{zhou2022learning} and CoCoOp~\cite{zhou2022conditional} both employ trainable prompts to improve zero-shot recognition performance, while TPT~\cite{shu2022test} leverages test-time adaptation of language prompts to improve out-of-distribution robustness. Visual prompting has also been proposed to reprogram recognition tasks~\cite{elsayed2018adversarial} and to enhance model performance on various downstream tasks~\cite{bahng2022exploring}. Recent research has also introduced memory-efficient prompt tuning purely for vision models~\cite{jia2022visual} to improve model generalization.
In this work, we propose visual prompt adaptation that achieves \textit{fully} test-time and \textit{storage-efficient} adaptation to improve model generalization.  

\noindent\textbf{Test-Time Adaptation}.
Adapting machine learning models to different test domains has been applied to numerous tasks~\cite{li2020model, liang2020we, kundu2020universal, sun2019test, sun2022benchmarking}. Among various adaptation techniques, we focus on test-time adaptation, which is particularly beneficial as it does not require label information from the test data~\cite{eastwood2021source, sun2019test, li2020model}. \textit{Source-free} adaptation allows models to adapt without any source-domain knowledge, adhering to real-world deployment constraints related to computation and privacy~\cite{kim2021domain, eastwood2021source}. \textit{Fully} test-time adaptation is more rigorous, as it necessitates on-the-fly model updates without delaying inference.\cite{schneider2020improving, nado2020evaluating} represent initial efforts towards achieving fully test-time adaptation, which involves updating or replacing the statistics of batch normalization (\texttt{BatchNorm}) layers\cite{ioffe2015batch} during inference. TENT accomplishes fully test-time adaptation by updating the model parameters in \texttt{BatchNorm} layers, using self-entropy minimization as its objective~\cite{wang2020tent}. MEMO takes advantage of input augmentations to achieve single image adaptation, circumventing the batch-level adaptation requirement in TENT~\cite{zhang2021memo}. In contrast to optimizing model parameters, an alternative approach is to adapt input with minor modifications. DDA employs diffusion models to purify input data, although it is limited to specific types of corruption~\cite{gao2022back}. TPT introduces test-time adaptation via language prompting to enhance the OOD robustness of the CLIP model~\cite{shu2022test}.

\noindent\textbf{Model Robustness against Distribution Shifts}. A trustworthy machine learning model should exhibit robust performance under data distribution shifts in real-world applications~\cite{sun2022spectral,sun2023pointdp}. Distribution shift refers to the differences between the underlying distributions of test and training data for a model trained on a specific dataset. Distribution shifts can naturally occur in the physical world due to environmental and conceptual variations~\cite{koh2021wilds,dhamija2018reducing}.In our study, we address three types of distribution shifts. Firstly, we consider natural variations such as object size, occlusion, and rendition changes as an OOD generalization problem, where the test set does not adhere to a specific pattern or concept. For instance, Hendrycks \textit{et al.} proposed ImageNet-A~\cite{hendrycks2021natural}, which serves as natural "adversarial" examples that challenge vision systems. ImageNet-R~\cite{hendrycks2021many} was introduced to encompass a variety of patterns, including art, cartoons, deviantart, and graffiti \textit{etc}. Secondly, we examine common corruptions of visual data that frequently occur in everyday life. For example, autonomous driving vehicles may encounter various weather changes, such as snow, fog, and rain~\cite{hendrycks2019benchmarking}. Lastly, we explore the domain adaptation problem, where the test set differs from the training set but follows a specific pattern or concept, such as DomainNet~\cite{kim2021domain} and VisDA-C~\cite{peng2017visda}. Numerous methods have been investigated to enhance the OOD robustness of machine learning models at different stages. These approaches include pretraining techniques~\cite{he2022masked,radford2021learning,zhou2021ibot}, finetuning methods~\cite{kumar2022fine,wortsman2022robust}, and test-time strategies~\cite{zhang2021memo,wang2020tent,gao2022back}. By exploring various techniques throughout the model's lifecycle, researchers aim to develop more robust models capable of handling distribution shifts problems. By tackling these distribution shifts, our study aims to improve the robustness and adaptability of machine learning models in real-world applications.

\section{VPA: Visual Prompt Adaptation}

In this section, we introduce VPA, which leverages visual prompting for fully test-time adaptation. We first motivate our design. Next, we describe our visual prompt and adaptation designs in \S\ref{sec:prompt_design} and \S\ref{sec:prompt_adaptation}, respectively. Finally, we present our adaptation setups in \S\ref{sec:adaptation_setting}.

\noindent\textbf{Why \textit{Fully} Test-Time?} Although various training-phase (\textit{i.e.}, pre-training and fine-tuning) methods have been proposed to improve model performance and generalization, there is no overarching combination that achieves the best performance. Therefore, test-time adaptation is a desirable complement. As briefly mentioned in \S\ref{sec:intro} and \S\ref{sec:related_work}, fully test-time adaptation updates the model without preventing inference nor accessing source domain information. Besides, it does not require any supporting dataset \cite{iwasawa2021test} and ensures that the adaptation solely relies on the current input (\textit{i.e.}, the batched or single image). We believe such a setup is realistic as it accounts for protecting the privacy and intellectual property of the model and supports domain switches during inference.

\noindent\textbf{Why Visual Prompting?} Visual prompting has shown great potential in adapting and even reprogramming the model during the training phase~\cite{chen2021exploring,jia2022visual}, making it memory and storage efficient. Visual prompt tuning is also beneficial in terms of faster convergence. Training a language model from scratch can be a time-consuming process; however, incorporating visual prompts can accelerate the training process by providing the model with relevant visual cues. This enables the model to learn and converge more quickly, resulting in improved efficiency and reduced training time. However, there are very few studies that have researched the application of visual prompting in test-time adaptation. Our study serves as a first step towards exploring the effectiveness of visual prompting in this area. Additionally, we integrate visual prompting into existing adaptation frameworks and demonstrate its superiority in improving model generalization and robustness.

\vspace{-0.1cm}
\subsection{Prompt Design}
\label{sec:prompt_design}

\noindent\textbf{Taxonomy of Visual Prompts}. In this study, we introduce a novel adaptation design that utilizes visual prompting. While prompting has been extensively studied in language models for various tasks, as discussed in \S~\ref{sec:related_work}, there exist only a few visual prompt designs aimed at improving recognition performance in the training phase~\cite{chen2021exploring,jia2022visual}. We formally define these designs as \textit{additive} and \textit{prependitive} prompts and illustrate them using the architecture of the Vision Transformer (ViT) model. Consider a ViT model with $N$ layers, where an input image is divided into $m$ patches $\{I_{i}| 1 \leq i \leq m \}$. Each patch is then fed into the linear projection layer with positional encoding: $\ve_0^{i} = \text{Linear}(I_{i})$. We denote the input to the $i$-th Transformer layer as $\mE_{i-1} = \{\ve_{i-1}^{j}|1 \leq j \leq m\}$, and the $i$-th Transformer layer output as $[CLS_i;\mE_i] = L_i([CLS_{i-1};\mE_{i-1}])$, where the classification head takes the final $CLS_{N}$ token for prediction: $y = \text{Head}(CLS_{N})$. Each Transformer layer consists of a self-attention module, an MLP layer with LayerNorm, and residual connections. The additive prompting is defined as:
\begin{equation}
\begin{split}
    [CLS_{i};\mE_i] = L_i([CLS_{i-1};\mP_{i-1}+\mE_{i-1}]) 
\end{split}    
\end{equation}
where $\mP_i=\{\vp_{i}^{j}| 1 \leq j \leq m\}$ and $+$ denotes element-wise addition.
Similarly, our prependitive prompting is formulated as:
\begin{equation}
    [CLS_{i};\mZ_i;\mE_i] = L_i([CLS_{i-1};\mP_{i-1};\mE_{i-1}]) 
\end{equation}
where the prompts $\mP$ play as additional tokens and $\mZ$ is the output corresponding to the input prompts. In the rest of this paper, we use $\oplus$ to denote the attachment of visual prompts \textit{i.e.}, $\mP \oplus \mE$. The numbers of visual prompts and layers prompted are configurable.

While we use the ViT architecture to demonstrate our visual prompt designs, it's important to note that additive prompting is compatible with most model architectures, as it only modifies the numeric values without altering the input size. In contrast, prependitive prompting changes the input dimension, which is better suited for ViT-based models, as Transformer blocks are insensitive to the length of the input~\cite{dosovitskiy2020image}.
In contrast to randomized initialization, we design a \textbf{zero attention} scheme to initialize the prompt with zero tensors. This approach ensures that the initialization process does not impact the original performance of the frozen model.

\begin{table*}[t]
\renewcommand\arraystretch{1}
\setlength\tabcolsep{8pt}
  \caption{OOD Generalization Evaluation Results (\%) of VPA on ImageNet Variants. The OOD average accuracy is calculated from the evaluation of ImageNet-A, ImageNet-R, and ObjectNet datasets.} 

  \vspace{-0.2cm}
  \label{tb:ood}
  \centering
  \resizebox{\linewidth}{!}{
    \begin{tabular}{l|l|cc|cccc}
\noalign{\global\arrayrulewidth1pt}\hline\noalign{\global\arrayrulewidth0.2pt}
Accuracy ($\uparrow$) & Method &ImageNet &ImageNet-V2 &ImageNet-A &ImageNet-R &ObjectNet &OOD Average\\
\hline
Source  & CLIP-ViT-LPFT &81.2 &71.1 &49.3 &71.1 &52.3 &57.6\\
\hline
\multirow{6}{*}{\makecell[c]{Episodic BIA}} 
 & TENT (Norm Layer) &81.3 &71.3 &49.6 &71.8 &52.6 &58.0\\
 & TENT (Cls Token) &81.2 &71.0 &49.4 &71.5 &52.3 &57.7\\
 & TENT (All Parameters) &81.2 &71.2 &49.7 &71.6 &52.4 &57.9\\
 & DDA  &77.2 &65.2 &38.5 &65.4 &46.5 &50.1\\
 
 & \textbf{Additiv VPA} &\textbf{81.3} &\textbf{71.4} &\textbf{50.4} &\textbf{72.0} &\textbf{52.8} &\textbf{58.4}\\
 & \textbf{Prependitive VPA} &\textbf{81.3} &71.3 &50.1 &\textbf{72.0} &52.5 &58.2\\
\hline
\multirow{3}{*}{\makecell[c]{Episodic SIA}} 
&MEMO & \textbf{81.3} & 72.3 &52.0 &72.2 &\textbf{52.9} &59.0\\
& \textbf{Additive VPA} &81.2 &72.3 &49.5 &\textbf{72.5} &52.3 &58.1\\
&  \textbf{Prependitive VPA} &81.2 &\textbf{72.9} &\textbf{52.4} &\textbf{72.6} &52.8 &\textbf{59.3}\\
\hline
\hline
Source & CLIP-ViT-WiSE &\textbf{79.8} &70.5 &49.7 &71.9 &52.4 &58.0\\
\hline
\multirow{6}{*}{\makecell[c]{Episodic BIA}} 
 & TENT (Norm Layer) &79.6 &70.7 &49.8 &72.2 &52.8 &58.3\\
 & TENT (Cls Token) &79.7 &70.5 &50.0 &72.2 &52.4 &58.2\\ 
 & TENT (All Parameters) &\textbf{79.8} &70.6 &50.3 &72.5 &52.5 &58.4\\
 & DDA &70.1 &62.2 &41.4 &64.8 &46.0 &50.7\\
 &  \textbf{Additive VPA} &\textbf{79.8} &\textbf{71.2} &\textbf{52.1} &\textbf{72.5} &\textbf{52.8} &\textbf{59.2} \\
 & \textbf{Prependitive VPA} &\textbf{79.8} &71.0 &51.2 &72.4 &52.5 &58.7\\
\hline
\multirow{3}{*}{\makecell[c]{Episodic SIA}} 
& MEMO &\textbf{80.1} &72.0 &53.9 &72.6 &53.0 &59.8\\
&  \textbf{Additive VPA} &80.0 &72.1 &50.3 &72.3 &52.5 & 58.4\\   
& \textbf{Prependitive VPA} &80.0 &\textbf{72.5} &\textbf{54.2} &\textbf{72.7} &\textbf{53.2} &\textbf{60.0} \\   
\hline
\hline

Source & CLIP-ResNet50$\times$4 &78.9 &67.5 &36.7 &64.0 &49.5 &50.1\\
\hline
\multirow{5}{*}{\makecell[c]{Episodic BIA}} 
 & BN &78.3 &67.5 &27.2 &55.1 &40.5 &40.9\\
 & TENT (Norm Layer) &78.1 &67.5 &27.3 &55.2 &40.4 &41.0\\
 & TENT (All Parameters) &79.1 &68.2 &37.2 &64.4 &49.9 &50.5\\
 & DDA  &69.0 &61.1 &24.6 &56.3 &41.0 &40.6\\
 & \ \textbf{Additive VPA} &\textbf{79.1} &\textbf{68.6} &\textbf{37.9} &\textbf{65.0} &\textbf{49.9} &\textbf{50.9} \\
\noalign{\global\arrayrulewidth1pt}\hline\noalign{\global\arrayrulewidth0.2pt}
\end{tabular}
  }
\vspace{-0.1cm}
\end{table*}

\vspace{-0.1cm}
\subsection{Prompt Adaptation}
\label{sec:prompt_adaptation}

In this section, we present the test-time adaptation procedure. Our study investigates two setups, namely \textit{episodic} and \textit{continual} adaptations. Episodic adaptation only applies to incoming data, and the model will be reset afterward. In contrast, continual adaptation lasts throughout the entire inference procedure. 

A recent study by Goyal et al.~\cite{goyal2022test} has demonstrated that self-entropy minimization is an almost-optimal objective function for episodic test-time adaptation on models trained with cross-entropy loss. Therefore, we adopt the objective of self-entropy minimization in our study. Let $f$ denote a well-trained classifier, and the self-entropy $H(\cdot)$ of a prediction is formulated as follows:
\begin{equation}
     H(\vz,\tau) = - \sum_{i=1}^{c} \sigma(\vz/\tau)_i \log \sigma(\vz/\tau)_i \quad
     \vz = f(\vx\oplus\mP)
\end{equation}
Here $c$ is the number of classes, $\sigma(\vz)_i = \frac{\exp{\vz(i)}}{\sum_{j=1}^{c}\exp{\vz}(j)}$ is the softmax function, and $\tau$ is a tunable hyper-parameter that controls the softmax temperature. Self-entropy is an unsupervised loss function as it relies only on predictions and not on ground-truth information. However, since entropy reflects the prediction confidence, it can serve as an indicator of the model's performance on the supervised task~\cite{wang2020tent}. As VPA is a general adaptation framework, we leverage both batched- and single-image adaptation settings in our work, as introduced below.

\noindent\textbf{Batched-Image Adaptation (BIA)}. Real-world machine learning services usually aggregate input data for batched predictions to save computation resources and time~\cite{batch}. Therefore, we mainly focus on BIA in our study, whose objective is formulated as:
\begin{equation}
\begin{split}
    \hat{\mP} = & \argmin_{\mP} \frac{1}{K}\sum_{i=1}^{K} H(f(\vx_i\oplus \mP),\tau) 
\end{split}
\end{equation}
where $K$ is the batch size. In BIA, VPA optimizes a visual prompt for all the test images in a given batch, which is the same setup as TENT~\cite{wang2020tent}, the SOTA method under BIA.

\noindent\textbf{Single-Image Adaptation (SIA)}. As self-entropy minimization requires batched inputs to function~\cite{wang2020tent}, we utilize the setups in MEMO~\cite{zhang2021memo} to expand a single image to a batch via augmentations. The objective of VPA under SIA is formulated as:
\begin{equation}
\begin{split}
    & \hat{\mP} = \argmin_{\mP} H(\frac{1}{\eta K}\sum_{i=1}^{K}f(S(\mathcal{A}_{i}(\vx_i),\eta)\oplus\mP,\tau) \\
    & S(\vx_i,\eta) =  \vx_i \cdot \mathbbm{1}[H(\vz_i,\tau) \leq \arg \text{top-}\eta K \{H(\vz,\tau)\}]
\end{split}
\end{equation}
where $\mathcal{A}$ denotes a random augmentation function, $K$ is the augmented batch size, and $S()$ is a confidence selection function to pick augmented images with high confidence with a percentile of $\eta$, following the setting in~\cite{shu2022test}. The intuition behind SIA is to use marginal entropy minimization on the augmented input to optimize the prompt and enhance generalization.

During both BIA and SIA, the visual prompt $\mP$ is optimized by computing the gradient of the entropy loss \textit{w.r.t.} $\mP$ (\textit{i.e.}, $\frac{\partial H}{\partial \mP}$) during the backward pass. As self-entropy only relies on the network's predictions without labels or source domain information, and our visual prompt $\mP$ is independent of model parameters, VPA achieves fully test-time adaptation. Additionally, besides episodic test-time adaptation, we also explore the application of continual online learning in test-time adaptation.

\noindent\textbf{Pseudo-Labeling Adaptation (PLA)}. Pseudo-labeling has been widely adopted in semi- and self-supervised learning. In this work, we adopt pseudo-labeling for test-time adaptation, which requires more setup than BIA and SIA. To implement this approach, we use a \textit{memory queue} $\mathcal{M}$ with size $s$ that stores the final $CLS_{N}$ token, along with the prediction of historic data $\vz_i$ for reference when processing an incoming batch, \textit{i.e.}, $\mathcal{M} = \{{CLS_{N}}_{i},\vz_i\}_1^{s}$. During adaptation, we generate reference labels for the incoming test data using its $k$ nearest-neighbor ($k$NN) predictions on the $CLS_{N}$ token before feeding into the head classifier. We then average the $k$NN predictions to produce the eventual pseudo label. We apply weak and strong augmentations to every incoming data sample inspired by FixMatch~\cite{sohn2020fixmatch}. Specifically, we obtain the soft predictions for the weakly and strongly augmented samples, denoted as $\vz_\mathcal{W}$ and $\vz_\mathcal{S}$, respectively. We then generate the pseudo label for the incoming data based on the soft voting mechanism in our memory queue:
\begin{equation}
    \hat{\vz_i} = \frac{1}{k} \sum_{j \in k\text{NN}_i} {\vz_\mathcal{W}}{j}
\end{equation}
We apply cross-entropy minimization as our objective, utilizing a temperature hyper-parameter on the generated soft pseudo label:
\begin{equation}
     H(\vz_\mathcal{S},\tau) = - \sum_{i=1}^{c} \sigma(\hat{\vz}/\tau)_i \log \sigma(\vz_\mathcal{S})_i \quad
     \vz = f(\vx\oplus\mP)
\end{equation}
\begin{equation}
    \hat{\mP} = \argmin_{\mP} H(f(\vx_i\oplus \mP),\tau) 
\end{equation}
Finally, we dynamically update the $CLS_{N}$ token along with its prediction $\vz_\mathcal{W}$ into our memory queue for next-round adaptation. Similarly, the visual prompt $\mP$ is optimized by computing the gradient of the cross-entropy loss.

\begin{table*}[t]
\renewcommand\arraystretch{1.1}
\setlength\tabcolsep{1pt}
  \caption{Corruption Robustness Evaluation Results (\%) of VPA on ImageNet-C with the Highest Severity Level.}
  \label{tb:eval_corrupt}
  \vspace{-0.2cm}
  \centering
  \resizebox{\linewidth}{!}{
  
    \begin{tabular}{l|l|ccccccccccccccc|c}
\noalign{\global\arrayrulewidth1pt}\hline\noalign{\global\arrayrulewidth0.2pt}
 Error Rate ($\downarrow$) &Method &Gauss. &Shot &Impulse &Defocus &Glass &Motion &Zoom &Snow &Frost &Fog &Bright &Contrast &Elastic &Pixelate &JPEG &Average\\
\hline
Source &  ViT & 52.5 & 51.7 & 51.5 & 56.0 &69.2 &51.0 &56.1 &46.3 &50.3 &45.6 &24.9 &69.1 &55.1 &34.2 &33.7 & 49.1\\
\hline
\multirow{2}{*}{\makecell[c]{Episodic SIA}} & MEMO & 48.5 & 47.8 &45.8 &58.2 &68.0 &53.5 &57.5 &42.7 &48.2 &41.8 &22.2 &68.9 &52.0 &32.1 &27.9 &47.7\\
&  \textbf{Prependitive VPA} & 48.2 &47.5 &45.4 &58.0 &67.6 &53.2 &56.0 &41.5 &48.3 &41.7 &22.5 &67.5 &52.2 &31.9 &27.5 &\textbf{47.3}\\
\hline
\multirow{3}{*}{\makecell[c]{Continual BIA}} & TENT (Norm Layer) & 47.7 & 49.0 & 45.5 &59.4 &68.1 &49.0 &55.9 &49.5 &48.6 &42.0 &22.3 &62.1 &52.1 &32.3 &27.5 &47.4\\
&  \textbf{Additive VPA} & 48.5 & 49.1 &45.7 &58.9 &67.5 &48.8 &55.8 &56.1 &49.7 &42.1 &22.1 &61.8 &52.5 &32.3 &27.7 &47.9\\
& \textbf{Prependitive VPA} &47.0 &48.5 &44.2 &56.8 &65.7 &48.2 &55.5 &48.2 &48.0 &40.1 &21.8 &61.4 &51.3 &31.2 &27.2 &\textbf{46.5}\\
\hline
\multirow{3}{*}{\makecell[c]{Continual PLA}} &  AdaContrast & 45.8 & 44.7 &44.5 &47.2 &57.8 &41.8 &46.0 &35.2 &39.8 &34.8 &22.8 &47.5 &40.2 &28.5 &29.5 &40.4\\
&  CFA &43.1 &42.0 &41.9 &45.6 &51.1 &40.1 &43.4 &33.6 &35.9 &32.3 &21.0 &41.2 &35.7 &28.3 &29.8 &37.6\\

 &   \textbf{Prependitive VPA} &46.7 &44.7 &43.9 &42.0 &44.5 &38.9 &43.0 &31.0 &33.2 &28.5 &22.9 &37.1 &31.8 &28.4 &30.0 &\textbf{36.6}\\
\noalign{\global\arrayrulewidth1pt}\hline\noalign{\global\arrayrulewidth0.2pt}

\end{tabular}}
\end{table*}

\subsection{Adaptation Setups}
\label{sec:adaptation_setting}

In our study, we primarily utilize the ViT-B/16 model architecture, which is now considered a standard benchmarking model~\cite{dosovitskiy2020image}. Additionally, we use ResNet~\cite{he2016deep} to demonstrate the general effectiveness of additive VPA. We attach additive prompts to the 1st and 6th Transformer layers in ViT, which are in total $196\times2 = 392$ tokens. Similarly. we insert 50 adaptable prompts into every other layer of the ViT-B architecture, resulting in 300 tokens, for prependitive prompting. By default, we set the value of $\tau$ to 1.0 in episodic adaptation, and we analyze its impact in Section~\ref{sec:eval_abla} to demonstrate that an optimal $\tau$ can lead to further improvements. We follow TENT~\cite{wang2020tent} and set 10 adaptation steps for each batch. We also ablate the number of steps in Section~\ref{sec:eval_abla}. We empirically set the learning rate to $4.0$ and $0.001$ with SGD~\cite{ruder2016overview} for additive and prependitive prompting, respectively. We use a batch size of 64 for all experiments in Section~\ref{sec:eval}. For SIA, we utilize random cropping~\cite{randomcrop} as the augmentation function $\mathcal{A}$ and $\mathcal{W}$. In the experimentation of PLA, we set $\tau = 0.07$ and the memory queue size to 1\% of the test dataset by default. We utilize $k=11$ similarly to AdaContrast~\cite{chen2022contrastive}. Moreover, we leverage random cropping and RandAugment~\cite{cubuk2020randaugment} as the weak ($\mathcal{W}$) and strong ($\mathcal{S}$) augmentations, respectively. 
Most importantly, we compare both \textit{episodic} and \textit{continual} learning settings in TENT~\cite{wang2020tent} with our VPA in this study.

\section{Experiments and Results}
\label{sec:eval}

This section reports on the experimental results of VPA and several other baseline methods across multiple benchmarks. As previously noted in \S~\ref{sec:intro}, we commence by assessing the performance of VPA in the presence of challenging distribution shifts (\textit{i.e.}, OOD generalization) in \S~\ref{sec:eval_ood}. Following this, we delve into exploring the potential of VPA in enhancing robustness against common corruptions in \S~\ref{sec:eval_corruption}. Lastly, we evaluate the performance of VPA on the domain adaptation task, alongside other baseline methods, in \S~\ref{sec:eval_da}.

\subsection{Evaluation of OOD Generalization}
\label{sec:eval_ood}

  


\noindent\textbf{Experimental Setups}. We select models pre-trained with CLIP~\cite{radford2021learning}, and leverage two SOTA robust fine-tuning methods (\textit{i.e.}, LPFT~\cite{kumar2022fine} and WiSE~\cite{wortsman2022robust}) to train them on the ImageNet training set.  To study model robustness to realistic OOD data that naturally occurs in the physical world, we leverage \textbf{ImageNet-A}~\cite{hendrycks2021natural}, \textbf{ImageNet-R}~\cite{hendrycks2021many}, and \textbf{ObjectNet}~\cite{NEURIPS2019_97af07a1}. ImageNet-A consists of 7,500 test images denoted as ``natural adversarial examples'' that are misclassified by a collection of standard models overlapped with 200 ImageNet categories. ImageNet-R collects 30,000 images of 200 ImageNet categories with artistic renditions. ObjectNet is a large real-world test set for object recognition with control where object backgrounds, rotations, and imaging viewpoints are random. We chose the ObjectNet subset that overlaps 113 classes with ImageNet in our study. We also use \textbf{ImageNet-V2}~\cite{recht2019imagenet}, a robustness benchmark with mild distribution shift, to further validate our results. As these challenging datasets do not follow specific distribution patterns, research has shown that continual learning may not be effective in improving robustness~\cite{chen2022contrastive,wang2020tent}. Therefore, we evaluate VPA under episodic BIA and SIA using different prompt types and measure the in-distribution (ID) and OOD accuracy as the metrics for our evaluation.

\vspace{-0.2cm}
\subsubsection{Evaluation on Foundation Models}
We present our large-scale evaluation of models fine-tuned with different methods in this study. Specifically, we utilize additive and prependitive prompting for VPA under BIA and SIA, respectively. As ViT models do not use \texttt{BatchNorm} layers, we default to adapting the \texttt{LayerNorm} layers for TENT. We also experiment with adapting the classification token $CLS_N$ and all model parameters for additional comparisons, following the settings in~\cite{kojima2022robustifying}. Table~\ref{tb:ood} shows the experimental results of different methods under BIA and SIA. We observe that TENT (Norm) only achieves slight improvements against natural distribution shifts compared to the source-only baseline. While the \texttt{LayerNorm} layer is a linear module that is preferred by the adaptation assumption in TENT~\cite{wang2020tent}, it is independent of the input data. Additionally, natural OOD data does not follow a clear distributional pattern, unlike synthesized corruptions. Therefore, the benefits of linear module adaptation do not transfer to our setting~\cite{wang2020tent}. Similarly, applying TENT to other model parameters does not show sensible improvements over the baseline either, which is consistent with prior studies~\cite{kojima2022robustifying}. In contrast, our VPA achieves visible enhancements over the source-only method. Specifically, VPA relatively improves OOD robustness by $3.8\%$ on average, while maintaining or improving the (near) ID accuracy on ImageNet and ImageNet-V2 for all chosen pre-trained models. Conversely, TENT degrades the original performance on (near) ID data in some cases. We also evaluate VPA and MEMO under SIA and find that both methods achieve around $2.7\%$ improvements over the source-only method (Table~\ref{tb:ood}). While MEMO adapts the whole backbone, VPA achieves more efficient prompt-level adaptation.
\begin{table*}[t]
\renewcommand\arraystretch{1.1}
\setlength\tabcolsep{6pt}
  \vspace{-0.2cm}
  \caption{Domain Adaptation Evaluation Results (\%) of VPA on DomainNet-126 (S: Sketch, R: Real, C: Clipart, I: Infograph, Q: Quickdraw, and P: Painting).}
  \label{tb:eval_da}
  \vspace{-0.2cm}
  \centering
  \resizebox{\linewidth}{!}{
  
    \begin{tabular}{l|l|cccccccccc|c}
\noalign{\global\arrayrulewidth1pt}\hline\noalign{\global\arrayrulewidth0.2pt}
Accuracy ($\uparrow$) & Method &S$\rightarrow$R &S$\rightarrow$C &S$\rightarrow$I &S$\rightarrow$Q &S$\rightarrow$P &R$\rightarrow$S &R$\rightarrow$C &R$\rightarrow$I &R$\rightarrow$Q &R$\rightarrow$P &Average\\

\hline
Source & CLIP-ViT-LPFT &67.5 &69.2 &35.5 &16.4 &52.8 &55.2 &68.6 &44.4 &9.7 &60.9 &48.0\\
\hline
\multirow{2}{*}{\makecell[c]{Episodic SIA}}
 & MEMO & 67.8 & 69.5 &35.4 &16.8 &53.2 &55.5 &68.9 &44.8 &10.1 &61.3 &48.3\\
&  \textbf{Prependitive VPA} &68.1 &70.1 &36.5 &17.3 &53.6 &55.8 &69.5 &45.0 &10.6 &61.5 &\textbf{48.8} \\
  \hline
\multirow{2}{*}{\makecell[c]{Continual BIA}}
 & TENT (Norm Layer) &67.7 &69.0 &35.8 &16.6 &53.0 &55.3 &68.9 &44.5 &9.9 &61.0 &48.2 \\
&  \textbf{Prependitive VPA} &68.7 &69.8 &36.3 &16.8 &53.2 &55.8 &69.0 &44.8 &10.1 &61.3 &\textbf{48.6} \\
  \hline
\multirow{3}{*}{\makecell[c]{Continual PLA}}
 & AdaContrast & 70.2 & 72.0 & 36.5 & 18.0 &54.2 &58.4 &71.8 &45.5 &10.5 &63.3 &50.0\\ 
& CFA & 69.8 &72.2 &35.9 &17.8 &54.4 &58.7 &71.6 &45.7 &10.3 &63.0 &49.9\\
 & \textbf{Prependitive VPA} &70.8 &73.0 &37.0 &17.8 &55.3 &58.8 &71.9 &46.4 &10.4 &63.8 &\textbf{50.5} \\

\noalign{\global\arrayrulewidth1pt}\hline\noalign{\global\arrayrulewidth0.2pt}

\end{tabular}}
\end{table*}

Our results show that, unlike under BIA where additive and prependitive prompting achieve similar adaptation performance, the prependitive VPA performs better under SIA. As discussed in \S~\ref{sec:prompt_adaptation} and illustrated in Figure~\ref{fig:overview}, SIA realizes test-time adaptation via augmentations over a single image. The workflow of SIA is as follows: 1) expand a single image as a mini-batch through augmentations for self-entropy minimization, 2) discard the expanded batch after adaptation, and 3) use the original single image and adapted model/prompt for inference. This setting is designed for \textit{model-level} adaptation, as the model is trained to be insensitive to input augmentations. However, \textit{additive prompts} are directly added on top of the input with a small magnitude, which makes their effectiveness sensitive to any change in the images. The prompt is adapted to the augmented batch but added to the original image in SIA, resulting in reduced effectiveness. On the other hand, prependitive prompting does not directly modify the semantics of the input, and the optimization over a single image is easier to converge than BIA with an appropriate prompt length. Our experiments show that prependitive prompting achieves a $3.3\%$ improvement under SIA, while additive prompting does not provide a tangible enhancement compared to the source-only baseline method.

We conduct another experiment of VPA on the ResNet architecture. Specifically, we use ResNet50x4 pre-trained with CLIP, and Table~\ref{tb:ood} presents the evaluation results, where VPA achieves similar improvement on the ResNet model~\cite{wortsman2022robust}. Surprisingly, we find that TENT has an around 20\% performance drop on ResNet50x4. TENT, by default, replaces the original statistics (\textit{i.e.}, $\mu$ and $\sigma$) in \texttt{BatchNorm} layers with the statistics of the input data. This setting is useful when the input batch is from one specific domain (\textit{e.g.}, synthesized corruptions). However, natural distribution shifts do not follow such assumptions, rendering significant performance degradation. In comparison, VPA consistently achieves better OOD robustness on different fine-tuning methods and model architectures.

\begin{figure*}
\vspace{-0.25cm}
  \subfigure[Different Number of Adaptation Steps.\label{fig:num_step}]{
    \includegraphics[width=0.48\linewidth, height=4cm]{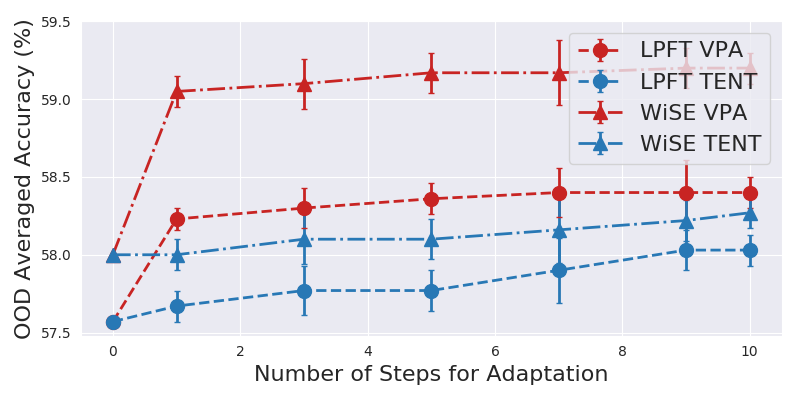}
    }
  \hfill
  \subfigure[Different Softmax Temperatures $\tau$.\label{fig:temperature}]{\includegraphics[width=0.48\linewidth, height=4cm]{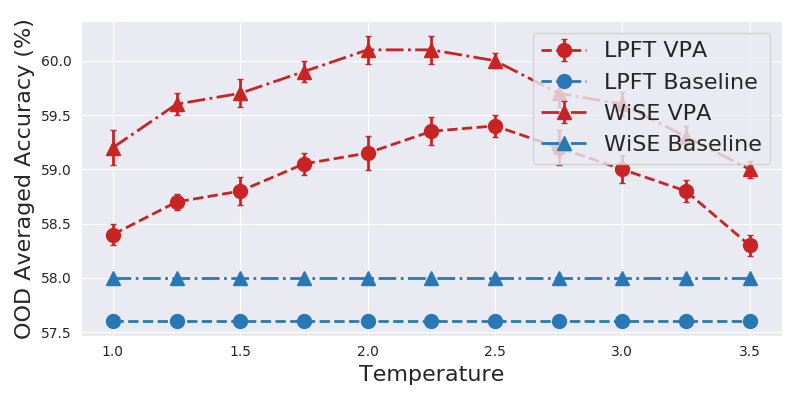}}
  \vspace{-0.3cm}
  \caption{Ablation Studies of Visual Prompt Adaptation on OOD Generalization. We show that the first adaptation step contributes the most in VPA and additional improvements can be achieved with an optimal temperature hyperparameter $\tau$.}
  \label{fig:short}
\end{figure*}

\subsection{Evaluation of Corruption Robustness}
\label{sec:eval_corruption}

\noindent\textbf{Experimental Setups}. 
In this section, we evaluate the performance of VPA against common corruptions using the \textbf{ImageNet-C} dataset~\cite{hendrycks2019benchmarking}. ImageNet-C is designed to assess the robustness and generalization capabilities of computer vision models by introducing 15 different corruptions at five severity levels to the original ImageNet validation dataset. These corruptions include various types of noise, blur, and distortion, making ImageNet-C a more realistic and challenging test of model robustness and generalization. We adopt the methodology from~\cite{chen2022contrastive}, focusing on the highest corruption severity level's performance. Our primary interest lies in the continual adaptation setting, as it has been shown to be more effective in enhancing corruption robustness; this is because each corruption can be considered as being drawn from a similar distribution shift. Based on our experiments, which demonstrated that the prependitive prompt outperforms the additive design under SIA, we employ only prependitive prompts for this evaluation. As foundation models do not show visible improvements on corruption robustness benchmarks, we use the ViT model pretrained on ImageNet in this section. In addition, we compare VPA with MEMO, a SOTA episodic adaptation method. Error rate serves as our evaluation metric for assessing corruption robustness.

Table~\ref{tb:eval_corrupt} presents the experimental results for the highest severity level, demonstrating that our prependitive VPA consistently achieves the best robustness improvement among all the baselines. Importantly, VPA attains the highest robustness under episodic SIA and provides greater storage efficiency compared to MEMO. This efficiency results from VPA adapting only the additional prompts, whereas MEMO adapts all the model parameters. For the evaluation under continual BIA, our prependitive VPA outperforms TENT and the additive design by 1.9\% and 3.0\%, respectively. This outcome may be associated with the nature of corruption benchmarks. Additive prompts are directly added to the embedding of the corrupted input, while prependitive prompts do not directly mix with the original corrupted embedding. Instead, they leverage the attention mechanism to interact with the original embedding, leading to a better robustness gain. Continual PLA achieves the largest gain, benefiting from the pseudo labels generated by the memory queue. Our prependitive VPA outperforms AdaContrast~\cite{chen2022contrastive} and CFA~\cite{kojima2022robustifying} by relative margins of 10.4\% and 2.7\%, respectively.

\subsection{Evaluation of Domain Adaptation}
\label{sec:eval_da}

\noindent\textbf{Experimental Setups}. In this section, we discuss our experiments and results related to the domain adaptation task. We employ the \textbf{DomainNet-126} dataset~\cite{peng2019moment} for this purpose. DomainNet encompasses common objects from six domains (\textit{i.e.,} sketch, real, clipart, infograph, quickdraw, and painting) and 345 categories. In our study, we empirically use the sketch (S) and real (R) images as the training sets and evaluate the adaptation performance on the remaining subsets. We leverage the ViT model pretrained with CLIP and finetuned by LPFT in this evaluation. For other setups, we follow the same configuration as used in the corruption robustness evaluation.

Table~\ref{tb:eval_da} displays the evaluation results for DomainNet-126, where we observe that VPA consistently achieves the best robustness across various adaptation settings. DomainNet shares certain similarities with ImageNet-C, as each domain exhibits a specific pattern. On average, our VPA outperforms the source-only baseline by 5.2\%. Furthermore, PLA-based VPA demonstrates more substantial improvements, with relative margins of 1.0\% and 1.1\% for AdaContrast and CFA, respectively. This highlights the effectiveness of our VPA approach in addressing domain adaptation challenges while maintaining robust performance across different settings.

\subsection{Ablation Studies}
\label{sec:eval_abla}

Besides visual prompt designs, this section provides an empirical analysis of VPA on different hyper-parameter settings under different adaptation settings. 

\noindent\textbf{Prompt Size}. We begin by conducting ablation studies on the prompt size utilized in additive and prependitive VPA. As outlined in \S~\ref{sec:adaptation_setting}, we utilize 50 adaptable prompts in 6 layers as the default for our prependitive VPA and 196 adaptable prompts in 2 layers of the ViT-B model for our additive VPA, leading to 300 and 392 learnable tokens, respectively. We vary the number of prompts adapted and assess the VPA's performance on OOD generalization and corruption robustness. Specifically, we vary the number of prompted layers for the additive VPA. The results are presented in Table~\ref{tb:abla}. The evaluation highlights the existence of an optimal point for the number of prompting tokens. Having too many prompts can make it difficult to optimize, while a relatively small number of prompts may restrict the capability of VPA.

\begin{table}[t]
    \centering
    \renewcommand\arraystretch{1}
    \setlength\tabcolsep{4pt}
    \caption{Ablation Study of the Prompt Size in Episodic VPA.}
    \label{tb:abla}
    \vspace{-0.2cm}
    \resizebox{\linewidth}{!}{
    \begin{tabular}{l|c|ccc}
    \noalign{\global\arrayrulewidth1pt}\hline\noalign{\global\arrayrulewidth0.2pt}
         Performance (\%)& Prompt Size & OOD ($\uparrow$) & Corruption ($\downarrow$) & DA ($\uparrow$)\\
        \hline
        \multirow{3}{*}{\makecell[c]{Additive VPA}} & 196 &58.5 &48.3 &-\\
        & 392  & \textbf{58.8} &\textbf{47.9} & -\\
        & 588 & 58.7 & 48.1 & -\\
        \hline
        \multirow{3}{*}{\makecell[c]{Prependitive VPA}} &150 &58.1 &47.0 &48.2\\
        & 300 &\textbf{58.5} &\textbf{46.5} &\textbf{48.6}\\
        & 450 &58.3 &46.8 &48.4\\
    \noalign{\global\arrayrulewidth1pt}\hline\noalign{\global\arrayrulewidth0.2pt}
    \end{tabular}
    }
\vspace{-0.4cm}
\end{table}

\noindent\textbf{Adaptation Steps and Temperature $\tau$}. We then ablate the effect of adaptation steps and the temperature parameter of additive episodic adaptation. In Figure~\ref{fig:num_step}, the average OOD accuracy generally improves as the number of steps increases. Encouragingly, we find the first step of VPA contributes the most to the OOD robustness improvement, where the relative improvements are $1.2\%$ and $1.9\%$ for models fine-tuned with LPFT and WiSE, respectively. In contrast, one-step TENT adaptation shows no improvements over the source-only baseline, demonstrating the effectiveness of VPA. Figure~\ref{fig:temperature} shows that there is a sweet point for the temperature parameter for OOD robustness improvement: We find that with an optimal temperature, the OOD robustness of the LPFT model could further improve by $1.5\%$ on average. However, selecting an optimal $\tau$ requires an additional validation set with access to the label, so we do not tune the temperature $\tau$ in the central part of our evaluation. We leave this as a future study to automatically select the temperature parameter for VPA. 

\noindent\textbf{Vision-Language Model}. We evaluate the performance of our proposed VPA combined with TPT~\cite{shu2022test} on the zero-shot recognition of the CLIP model. We adopt the SIA experimental setup from~\cite{shu2022test}. The results in Table~\ref{tab:vl} demonstrate that our VPA consistently enhances the zero-shot recognition performance on the challenging OOD generalization benchmarks by approximately 0.5\%.

\begin{table}[t]
    \centering
        \vspace{-0.2cm}
    \setlength\tabcolsep{2pt}
    \caption{OOD Generalization Evaluation Results (\%) of VPA for Zero-Shot Recognition in the Vision-Language Model.}
    \label{tab:vl}
    \resizebox{\linewidth}{!}{
    \begin{tabular}{l|cc|cc}
    \noalign{\global\arrayrulewidth1pt}\hline\noalign{\global\arrayrulewidth0.2pt}
        & ImageNet & ImageNet-V2 & ImageNet-A & ImageNet-R\\
        \hline
        CLIP & 66.7 &60.9 &47.9 &74.0 \\
        CLIP+TPT &69.0 &63.4 &53.5 &76.5 \\
        CLIP+TPT+\textbf{VPA} &\textbf{69.1} &\textbf{63.7} &\textbf{53.9} &\textbf{77.0} \\
        \noalign{\global\arrayrulewidth1pt}\hline\noalign{\global\arrayrulewidth0.2pt}
    \end{tabular}
    }
\end{table}


\section{Concluding Remarks}
To conclude, we propose VPA, a pioneering framework for generalizing visual prompting with test-time adaptation. VPA effectively improves OOD generalization, corruption robustness, and domain adaptation performance across diverse settings and tasks. The effectiveness of VPA highlights the potential of incorporating visual prompting in future research to address a wide array of adaptation challenges.



\begin{acks}
We appreciate the constructive feedback from the anonymous reviewers. Jiachen Sun and Z. Morley Mao were partially supported by NSF under CNS-1930041, CMMI-2038215, and the National AI Institute for Edge Computing Leveraging Next Generation Wireless Networks, Grant \# 2112562.
\end{acks}

\clearpage
\clearpage

{
\bibliographystyle{ACM-Reference-Format}
\balance
\bibliography{sample-base}
}

\clearpage
\appendix
\section*{Appendices}

\section{Discussion}
In this paper, we have investigated the application of visual prompting in fully online test-time adaptation. Although prompting has been extensively studied in NLP tasks and has recently gained significant attention for improving the zero-shot performance of vision-language models, its exploration in vision systems remains limited. In NLP tasks, prompted embedding tends to lose semantics after tuning, unlike na"ive prompt engineering~\cite{zhou2022learning}. Similarly, explaining the operational principle of visual prompting remains challenging. As introduced in~\cite{elsayed2018adversarial}, a more general interpretation of visual prompting involves reprogramming a well-trained vision model to achieve any deterministic goal. Despite the inherent limitations of fully test-time visual prompting, we have explored its application across three critical aspects of real-world machine learning systems: OOD generalization, corruption robustness, and domain adaptation, demonstrating its effectiveness. Our experimental results generally show that prependitive VPA is more effective, whereas additive VPA is more universal to different model architectures. We believe visual prompting could also be applied to other tasks as ViT architectures advance and become more dominant. Another avenue for future research is prompt design. In this work, we have explored four combinations of prompting setups in \S~\ref{sec:prompt_design}; additional design choices, including image-to-image models~\cite{zhu2017unpaired} and embedding-space prompting~\cite{jia2022visual}, could yield intriguing applications in the future. It is important to note that our study's focus is not solely on achieving state-of-the-art results but rather on examining how visual prompting performs within the fully test-time adaptation framework. We are aware of a concurrent, yet unpublished work, DePT~\cite{gao2022visual}, which investigates a similar topic. DePT primarily focuses on offline test-time adaptation, where the model is adapted offline. Offline adaptation offers more room for sophisticated prompting designs. We also noticed another concurrent work~\cite{gan2023decorate} that studies a similar problem. We believe that all three studies are complementary, each providing valuable insights into different aspects of test-time adaptation using visual prompting.

\section{Ablation Studies}
We further conducted two ablation studies on the augmentation and $k$NN soft majority voting in PLA.

\begin{table}[h]
    \centering
        \vspace{-0.2cm}
    \setlength\tabcolsep{8pt}
    \caption{Ablation Study on Augmentation Method in VPA.}
    \label{tab:ablate_1}
    \resizebox{\linewidth}{!}{
    \begin{tabular}{l|cc}
    \noalign{\global\arrayrulewidth1pt}\hline\noalign{\global\arrayrulewidth0.2pt}
        PLA (\%) & ImageNet-C ($\downarrow$) & DomainNet-126 ($\uparrow$) \\
        \hline
        RandAugment & 36.6 &50.5\\
        \rowcolor{gray!20}
        AugMix &\textbf{35.7} &\textbf{51.1} \\
        \noalign{\global\arrayrulewidth1pt}\hline\noalign{\global\arrayrulewidth0.2pt}
    \end{tabular}
    }
\end{table}

In single-image adaptation (SIA), we employ random cropping as the preferred augmentation technique. The weak augmentation employed in pseudo-label adaptation also utilizes random cropping, whereas the strong augmentation is constructed based on RandAugment~\cite{cubuk2020randaugment}. We here utilize AugMix~\cite{hendrycks2019augmix} to evaluate the impact of distinct augmentation techniques. Our findings reveal that the usage of AugMix can further boost the performance of Visual Prompting Adaptation (VPA) within the pseudo-label adaptation framework by approximately 0.8\%, as shown in Table~\ref{tab:ablate_1}. These findings illustrate the considerable potential for performance enhancement through the thoughtful selection and application of image augmentation techniques.

We default to leverage $k = 11$ in the soft majority voting. We further ablate the importance of $k$ in this experiment. As presented in Table~\ref{tab:ablate_2}, $k$ is indeed an essential hyper-parameter in pseudo-label adaptation. We find that $k \in [11,15]$ generally achieves the highest performance gain.

\begin{table}[h]
    \centering
        \vspace{-0.2cm}
    \setlength\tabcolsep{8pt}
    \caption{Ablation Study on $k$NN Soft Majority Voting.}
    \label{tab:ablate_2}
    \resizebox{0.6\linewidth}{!}{
    \begin{tabular}{l|c}
    \noalign{\global\arrayrulewidth1pt}\hline\noalign{\global\arrayrulewidth0.2pt}
        PLA (\%) & ImageNet-C ($\downarrow$) \\
        \hline
        $k=3$ & 44.1\\
        $k=7$ & 39.0\\
        $k=11$ & 36.6\\
        \rowcolor{gray!20}
        $k=15$ & \textbf{36.4}\\
        $k=21$ & 38.2\\
        \noalign{\global\arrayrulewidth1pt}\hline\noalign{\global\arrayrulewidth0.2pt}
    \end{tabular}
    }
\end{table}

\end{document}